# MULTIMODAL FUSION VIA A SERIES OF TRANSFERS FOR NOISE REMOVAL

*Chang-Hwan Son and  Xiao-Ping Zhang*

Department of Electrical and Computer Engineering, Ryerson University

## ABSTRACT

Near-infrared imaging has been considered as a solution to provide high quality photographs in dim lighting conditions. This imaging system captures two types of multimodal images: one is near-infrared gray image (NGI) and the other is the visible color image (VCI). NGI is noise-free but it is grayscale, whereas the VCI has colors but it contains noise. Moreover, there exist serious edge and brightness discrepancies between NGI and VCI. To deal with this problem, a new transfer-based fusion method is proposed for noise removal. Different from conventional fusion approaches, the proposed method conducts a series of transfers: contrast, detail, and color transfers. First, the proposed contrast and detail transfers aim at solving the serious discrepancy problem, thereby creating a new noise-free and detail-preserving NGI. Second, the proposed color transfer models the unknown colors from the denoised VCI via a linear transform, and then transfers natural-looking colors into the newly generated NGI. Experimental results show that the proposed transfer-based fusion method is highly successful in solving the discrepancy problem, thereby describing edges and textures clearly as well as removing noise completely on the fused images. Most of all, the proposed method is superior to conventional fusion methods and guided filtering, and even the state-of-the-art fusion methods based on scale map and layer decomposition.

## 1. INTRODUCTION

Recently, near-infrared imaging [1] has introduced to provide the noise-free Near-infrared Gray Image (NGI) and the corresponding noise-contained Visible Color Image (VCI), which are captured consecutively for the same scene during a shoot in dim lighting conditions. This near-infrared imaging system can be realized by using near-infrared pass-/block-filters [1] or near-infrared flash [2]. Since the captured NGI is noise-free, the near-infrared imaging has been considered as a solution for high-quality image acquisition in low lighting conditions.

### 1.1. Related works

Conventional near-infrared fusion models based on the gradient difference regularization (GDR) [3-5], multiresolution (MR) [6-8], and, weighted least squares (WLS) [2] fail to remove noise and to describe details. The first reason is that the conventional fusion models are not suitable for overcoming the serious discrepancies in the edges and brightness between NGI and VCI. The captured NGI and VCI have different characteristics regarding color and noise. More specifically, the NGI is noise-free but it is grayscale, whereas the VCI has colors but it contains noise. This different modality can be checked in the Fig. 1 where edge and brightness discrepancies exist around the red line of the bower and petal, respectively. The second reason is that the conventional methods follow the traditional fusion strategy, which attempts to combine the NGI with the VCI based on regularization and multiresolution fusion models (e.g., GDR, MR, and WLS) in the transform domain or spatial domain. In other words, conventional fusion methods can be considered as the weighted averaging of two input images even though the NGI is used as the guidance image with the fusion models. Therefore, clean background and sharp edges of the NGI are inevitably fused with the noise of the VCI during multimodal image fusion. Recently-introduced guided filtering (GF) [9] differs from the traditional fusion strategy, however color distortion appears on the resulting images, due to the brightness discrepancy. This reveals that GF is closer to edge-preserving smoothing filters (e.g., bilateral filter), which were not designed to overcome the discrepancy problem. Naturally, the GF-based image fusion [10] for high dynamic rendering (HDR) is not suitable for the near-infrared fusion. More recently, the state-of-the-art methods based on scale map (SM) [11] and layer decomposition (LD) [12] are introduced to handle the serious discrepancy problem. However, SM method tends to remove textures and over-enhance original colors. Even though the LD method is much stronger for the detail description than the conventional fusion and GF methods, there is still room for improvement in noise removal. Moreover, LD method is too slow, due to the time-consuming sparse coding.

## 2. FUSION STATEGY

Figures 1 and 2 show how the proposed method is quite different from the conventional fusion methods for noise removal. The conventional fusion methods transform multimodal images into the transform domain (e.g., Wavelet/FFT domains) to utilize the edge strength and entropy ratios between the noise and clean images at different scales [6-8,13], or to provide the fast computation of the guided filtering based on the GDR [3-5]. In addition, spatially varying norms can be used, according to the gradient distributions (e.g., flat/texture/edge distributions) [14]. In the spatial domain, regularization parameter can be adjusted adaptively, according to the edges of the guidance image with the WLS model [2].

### 2.1. Our fusion strategy

The proposed method conducts a series of transfers: contrast, detail, and color transfers, as shown in Fig. 2. First, the purpose of the contrast and detail transfers is to solve the discrepancy problem in the edge and brightness between the VCI and the NGI, and then generate a new NGI with more improved edges and brightness. As shown in Fig. 2, the newly created NGI is totally different from the captured NGI. Specifically, the line near the brim of the bower is restored more clearly on the newly created NGI and the brightness of the flower is also corrected to be more realistic. Second, the role of the color transfer is to correct the colors of the denoised VCI via a linear transfer model, and then add the corrected colors to the newly generated NGI. Note that the captured NGI is noise-free, and thus the newly generated NGI is also noise-free. Therefore, if we can generate colors to be added to the newly generated NGI, noise-free and edge-preserving high-quality photographs can be obtained. Before doing a series of transfers, initial denoising is required, as shown in Fig. 2, to completely remove the noise on the VCI, which ensures that the proposed contrast, detail, and color transfers work properly. In other words, if the initial denoising is excluded, the noise contained in the VCI can be transferred into the newly generated NGI, and thus the noise appears on the fused images as well.

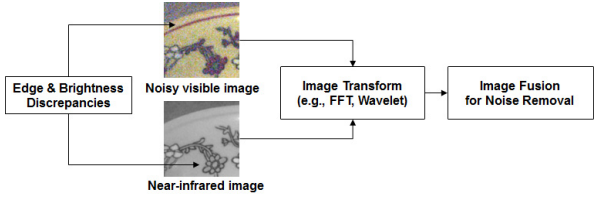

Fig. 1. Conventional fusion strategy for noise removal.

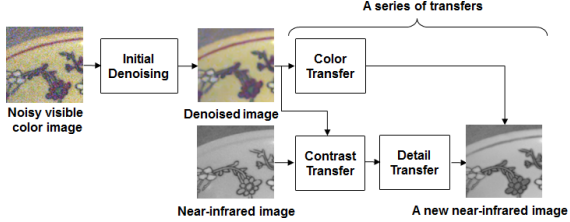

Fig. 2. Proposed transfer-based fusion strategy for noise removal.

In short, the proposed approach is based on a series of transfers. Different from conventional approaches, the proposed method transfers the modified colors of the denoised VCI into the newly generated NGI, not the captured NGI, and thus the proposed fusion method is closer to the transfer method [15]. This is the main difference between the proposed method and the conventional fusion methods such as GDR [3-5], MR [6-8], WLS [2], GF [9].

## 3. NOTATION

In this paper, bold lowercase is used to indicate column vectors. For example, $\mathbf{x}^v$ and $\mathbf{x}^n$ indicate the column vectors that contain the pixel values of the denoised VCI and NGI, respectively. The superscripts are used to differentiate between the two images. If the captured images are not grayscale, i.e., there are color channels, the superscript $c$ is additionally used to indicate the color channel like this $\mathbf{x}^{v(c)}$. Here, the superscript $c$ can be one of $\ell^*$, $\alpha^*$, or $\beta^*$, which indicate the luminance and chrominance planes of the decorrelated color space [16] or opponent color space [17]. To denote the pixel location, the subscript $i$ is used like $\mathbf{x}_i^v$.

## 4. DETAILS OF PROPOSED TRANSFERS

Before doing a series of transfers, initial denoising is required. It is important to completely remove the noise on the VCI during initial denoising to guarantee that the proposed contrast, detail, and color transfers can generate a noise-free NGI with more improved edges and brightness, and then add colors to the newly generated NGI. In this paper, non-local means filtering [18] is used for initial denoising. Certainly, other filtering methods (e.g., bilateral filtering [19]) can be considered.

### 4.1. Contrast transfer

The purpose of the contrast transfer is to solve the discrepancies in the edges and brightness between the denoised VCI and the NGI, and then generate another NGI with more improved edges and brightness, as illustrated in Fig. 2. The proposed color transfer is modeled as follows.

$$\min_{\mathbf{s}_i} \left\| \mathbf{W}_i^{1/2} (\mathbf{p}_i - \mathbf{Q}_i \mathbf{s}_i) \right\|_2^2 + \mu_c \left\| \mathbf{s}_i - \mathbf{s}_i^0 \right\|_2^2 \qquad (1)$$

where $\left\| \ \right\|_p$ indicates $p$-norm; $\mathbf{p}_i$ and $\mathbf{Q}_i$ contain the pixel information of the extracted patches from the luminance plane of the denoised VCI and NGI at the $i$th pixel location, respectively. Assuming that the extracted patch has an odd size $m \times m$ (e.g., $5 \times 5$), $\mathbf{p}_i$ and $\mathbf{Q}_i$ are given as follows,

$$\mathbf{p}_i = \mathbf{R}_i \mathbf{x}^{v(\ast)} \quad \text{and} \quad \mathbf{Q}_i = [\mathbf{R}_i \mathbf{x}^n \ \ \mathbf{1}] \qquad (2)$$

where $\mathbf{R}_i$ is a matrix that extracts the patch from an image at the $i$th pixel location [20]. If $\mathbf{x}^{v(\ast)}$ and $\mathbf{x}^n$ are image vectors with the size of $N \times 1$, the matrix $\mathbf{R}_i$ has dimensions $m^2 \times N$ . In (2), $\mathbf{1}$ indicates the column vector with all ones. Thus, the dimensions of $\mathbf{Q}_i$ and $\mathbf{p}_i$ are $m^2 \times 2$ and $m^2 \times 1$, respectively. In (1), $\mathbf{W}$ is a diagonal matrix including the weighting values that are inversely proportional to the distance between the center pixel location, $i$ , and its neighbor pixel location. $\mathbf{s}_i^T = [\mathbf{s}_i(1), \mathbf{s}_i(2)]$ contains two vector elements that indicate the slope and bias, respectively. Here, $T$ denotes the transpose operator. Therefore, the first term $\left\| \mathbf{W}_i^{1/2} (\mathbf{p}_i - \mathbf{Q}_i \mathbf{s}_i) \right\|_2^2$ can be regarded as a linear transform. In other words, the near-infrared patch $\mathbf{R}_i \mathbf{x}^n$ will be mapped into the visible luminance patch $\mathbf{R}_i \mathbf{x}^{v(\ast)}$ without any constraints. To prevent this, the local contrast-preserving regularization $\left\| \mathbf{s}_i - \mathbf{s}_i^0 \right\|_2^2$ is additionally used. This second term forces the unknown vector $\mathbf{s}_i$ to be close to $\mathbf{s}_i^0$, which is given via the following equation,

$$\omega_1 \left[ \mathbf{x}_i^n / avg(\mathbf{R}_i \mathbf{x}^n), \ 0 \right]^T + \omega_2 \left[ \mathbf{x}^{v(\ast)} / avg(\mathbf{R}_i \mathbf{x}^{v(\ast)}), \ 0 \right]^T \qquad (3)$$

where $avg$ is the averaging function. This equation indicates that the center-pixel values of the near-infrared patch and the denoised visible-luminance patch are divided by their respective average values, and then combined linearly with the weighting values $\omega$ , which are set by the variance ratio between the two patches. The ratio of the center-pixel brightness to the background brightness, as shown in (3), has been used as the contrast measure [21]. Note that the slope of $\mathbf{s}_i(1)$ corresponds to a local contrast in an image.

By adopting this local contrast regularization, the newly created near-infrared patch can have more improved contrast and edges. In other words, the edge and brightness discrepancies can be reduced.

The equation (1) is the convex function, and thus there exists a closed-form solution:

$$\mathbf{s}_i = (\mathbf{Q}_i^T \mathbf{W}_i \mathbf{Q}_i + \mu_c \mathbf{I})^{-1} (\mathbf{Q}_i^T \mathbf{W}_i \mathbf{p}_i + \mu_c \mathbf{s}_i^0) \qquad (4)$$

where $\mathbf{I}$ is the identity matrix. Given the estimated $\mathbf{s}_i$, the newly created NGI ( $\mathbf{x}^{o(i*)}$ ) is given, according to the equation: $\mathbf{x}_i^{o(i*)} = \mathbf{x}_i^n \mathbf{s}_i(1) + \mathbf{s}_i(2)$ .

## 4.2. Detail Transfer

The proposed contrast transfer can generate the noise-free NGI with more improved edges and brightness. However, fine details can be lost during the contrast transfer. To handle this issue, the detail layer of the newly generated NGI is modified, as follows:

$$\Delta \mathbf{x}^{o(i*)} = \mathbf{x}^{o(i*)} - \mathbf{x}^{o,b} \quad \text{and} \quad \Delta \mathbf{x}^n = \mathbf{x}^n - \mathbf{x}^{n,b} \qquad (5)$$

$$\min_{\Delta \mathbf{x}} \left\{ \mu_d \left\| \Delta \mathbf{x} - \Delta \mathbf{x}^{o(i*)} \right\|_2^2 + \sum_{j=1}^{2} \left| (\Delta \mathbf{x} \otimes f^j) - (\Delta \mathbf{x}^n \otimes f^j) \right| \right\} \qquad (6)$$

$$\mathbf{x}^{o(i*)} \leftarrow \mathbf{x}^{o,b} + \Delta \mathbf{x} \qquad (7)$$

where $\mathbf{x}^{o,b}$ and $\mathbf{x}^{n,b}$ denote the base layers of the newly created NGI and the captured NGI, respectively, and $\Delta \mathbf{x}^{o(i*)}$ and $\Delta \mathbf{x}^n$ indicate the corresponding detail layers. As shown in (7), the detail layer $\Delta \mathbf{x}$ is newly estimated via (6), and then added to the base layer $\mathbf{x}^{o,b}$ , thereby producing the detail-enhanced version $\mathbf{x}^{o(i*)}$ . In (6), the second term forces the gradient difference between the two detail layers to be small, and thus the details of the newly created NGI can be strengthened. In (5), nonlocal means filtering is used to generate the two base layers. Equation (6) can be solved by using the alternating minimization [22].

## 4.3. Color Transfer

In this section, we introduce how to transfer the colors of the denoised VCI into the newly created NGI ( $\mathbf{x}^{o(i*)}$ ). The mapping relation has already been established between the denoised VCI and the NGI during the contrast transfer. Thus, the unknown colors for the newly created NGI can be derived, as follows:

$$\mathbf{x}_i^{o(\alpha*)} = \mathbf{x}_i^{\nu(\alpha*)} / \mathbf{s}_i(1) \quad \text{and} \quad \mathbf{x}_i^{o(\beta*)} = \mathbf{x}_i^{\nu(\beta*)} / \mathbf{s}_i(1) \qquad (8)$$

where $\mathbf{x}^{\nu(\alpha*)}$ and $\mathbf{x}^{\nu(\beta*)}$ indicate the two chrominance planes of the denoised VCI. Above equation shows that the unknown chrominance planes $\mathbf{x}^{o(\alpha*)}$ and $\mathbf{x}^{o(\beta*)}$ corresponding to the newly created NGI $\mathbf{x}^{o(i*)}$ is obtained by dividing the chrominance planes of the denoised VCI by $\mathbf{s}_i(1)$ , which is the mapping relation. In other words, the equation (8) is derived from the linear mapping in (1), which is roughly expressed by $\mathbf{x}_i^n \mathbf{s}_i(1) \approx \mathbf{x}_i^v$ . This reveals that the unknown chrominance planes of the NGI can be defined as the contrast-enhanced version of the chrominance planes of the denoised VCI. By combining the newly created NGI $\mathbf{x}^{o(i*)}$ with its estimated chrominance planes $\mathbf{x}^{o(\alpha*)}$ and $\mathbf{x}^{o(\beta*)}$ in the decorrelated color space [16], the proposed method based on a series of transfers can produce noise-free and detail-preserving images.

## 5. ERIMENTAL RESULTS

Fig. 3 shows the experimental results for multimodal image fusion. Figs. 3(a) and (b) shows captured VCI and NGI in a dim lighting condition [2,11]. It is clear that edge and brightness discrepancy problems between the VCI and NGI are serious, especially for the chart' patches and red lines near the brim of the bowl. Fig. 3(c) shows the initially denoised VCI via nonlocal means filtering [18]. It is noticed that the noise is clearly removed. The boundary of the patches and lines are blurred, as well. However, this edge loss can be restored by applying the proposed contrast and detail transfers, as shown in Figs. 3(d) and (e). In Fig. 3(d), it is shown that the discrepancy problem is successfully solved via the proposed contrast transfer. For example, the red line on the bowl that was almost removed in the captured NGI of Fig. 3(b) is clearly restored. Also, the brightness ratios between the patches on the chart are corrected to be more realistic. In Fig. 3(e), the use of the detail transfer leads to an enhancement in the detail description. In Fig. 3(f) shows the fused color image via the proposed color transfer. Compared to Figs. (g)-(l), the proposed method is superior to the conventional fusion methods. This is thanks to the transfer-based fusion strategy, as discussed in Section 2. In other words, the proposed method transfers the corrected colors of the denoised VCI into the noise-free and edge-preserving NGI, as shown in Fig. 3(c). Therefore, there is no noise in the fused image of Fig. 3(f). Moreover, edges and background textures are preserved clearly.

BM3D [23] is known as one of the state-of-the-art denoising methods. However, its visual quality is poor in terms of edges and textures, as shown in Fig. 3(g). It is guessed that the main reason is due to the non-Gaussian noise in the captured VCI. The traditional fusion methods based on the WLS [2] and GDR [3-5] can produce better-resulting images than the BM3D, as shown in Figs. 3(h) and (i). This is thanks to the use of the noise-free NGI, which is used as a guidance image [3-5]. Fig. 3(j) shows the fused image with the GF method [9]. It is noticed that the visual quality level is low. This indicates that the GF is not suitable for solving the discrepancy problems. Figs. 3(k) and (l) show the fused images with the SM [11] and LD [12] methods, respectively, which are the state-of-the-art fusion methods. Nevertheless, the edges of the brim of the bowl are broken and the background colors are distorted with the SM method, as shown in Fig. 3(k). Even though the LD method is strong at the texture representation, the overall sharpness needs to be improved, compared to the proposed method. In addition, the LD method is too slow, due to the repetition of the residual sparse coding. Similar effects can be found for another captured image. As shown in Fig. 4, the proposed method produces more clear texts and background texture than the conventional fusion methods: GF, GDR, BM3D, SM, and LD methods.

## CONCLUSION

This paper introduces a new transfer-based near-infrared fusion method for noise removal. Different from the traditional approaches, the proposed method conducts a series of transfers. The experimental results showed that the proposed contrast and detail transfers are successful at solving the serious discrepancy problems between the captured VCI and NGI, thereby producing the noise-free and detail-preserving NGI. The subsequent color transfer produces the realistic colors by correcting the colors of the denoised VCI via linear transform model. Most of all, the proposed method is superior to conventional fusion methods.

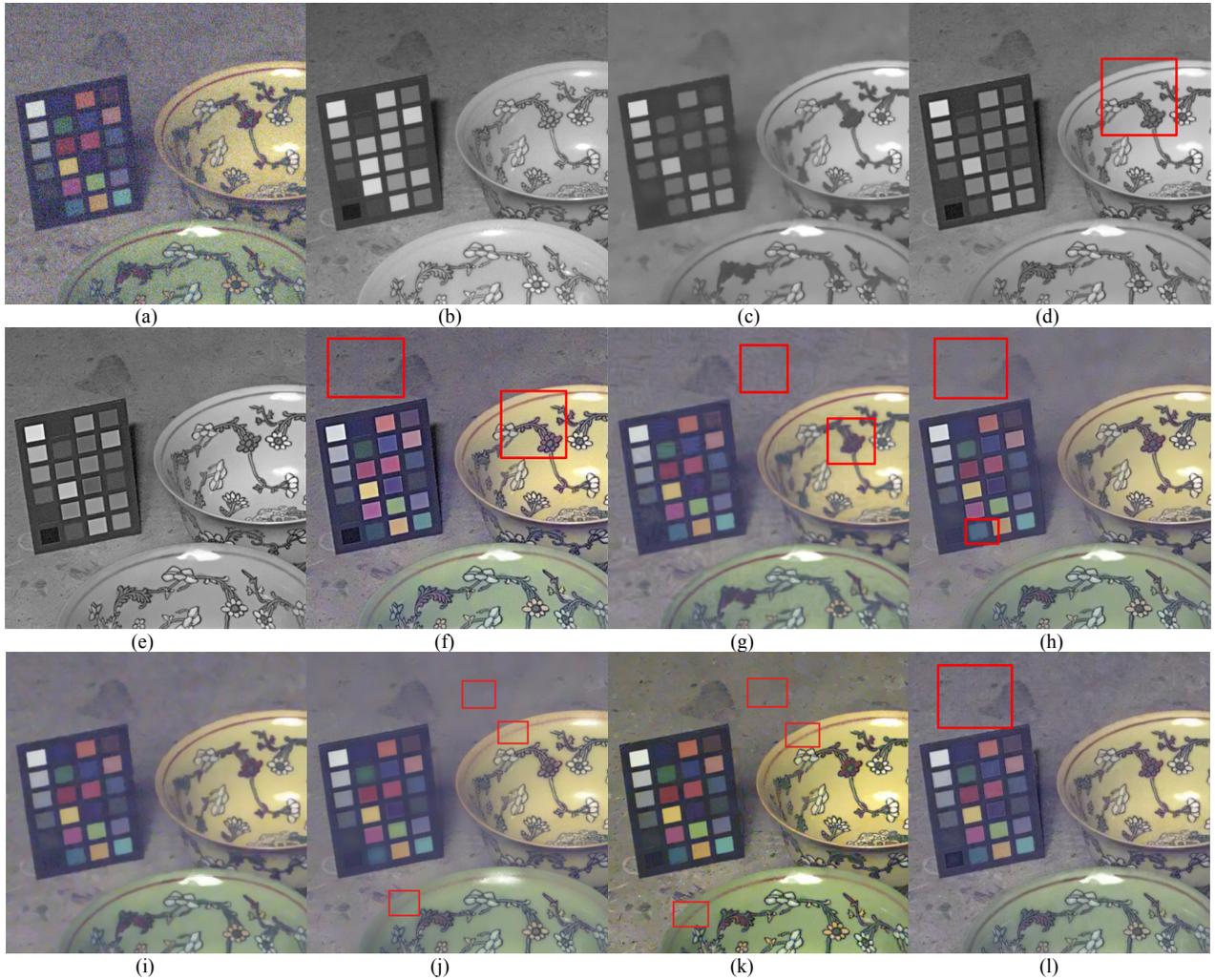

Fig. 3. Experimental results: (a) captured VCI, (b) captured NGI (c) initially denoised VCI via nonlocal means filtering [18], (d) newly created NGI via the proposed contrast transfer, (e) newly created NGI via the detail transfer, **(f) fused image with the proposed color transfer**, (g) BM3D [23], (h) GDR [3], (i) WLS [2], (j) GF [9], (k) SM [11], and (l) LD [12].

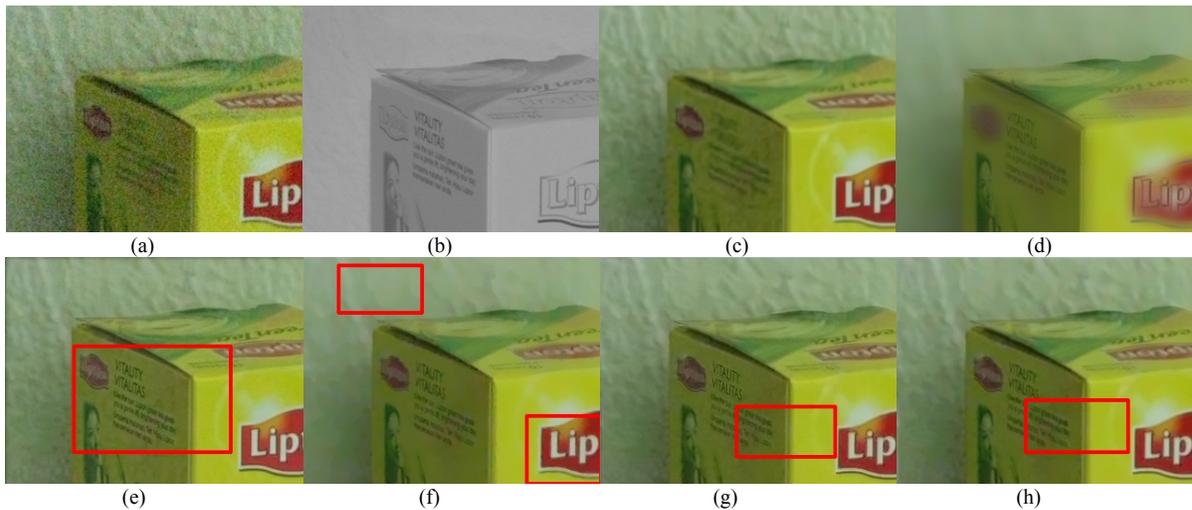

Fig. 4. Experimental results: (a) noisy VCI, (b) NGI (c) BM3D [23], (d) GF [9], (e) **proposed meethod**, (f) SM [11], (g) LD [12], and (h) GDR [3].